\documentclass[runningheads,a4paper]{llncs}

\usepackage{amssymb}
\usepackage{graphicx}
\usepackage{amsmath,graphicx}
\usepackage{multirow}
\usepackage{url}
\usepackage{array}
\usepackage{geometry}
\urldef{\mailsa}\path|{yuying5,faraji}@ualberta.ca|
\newcommand{\keywords}[1]{\par\addvspace\baselineskip
\noindent\keywordname\enspace\ignorespaces#1}

\begin{document}

\mainmatter  

\title{EREL Selection using Morphological Relation}

\titlerunning{EREL Selection using Morphological Relation}

%
%
\author{Yuying Li \and Mehdi Faraji}
\authorrunning{Yuying Li \and Mehdi Faraji}

\institute{Department of Computing Science,\\
University of Alberta, Canada\\
\mailsa\\
}

%
%

\toctitle{Lecture Notes in Computer Science}
\tocauthor{Authors' Instructions}
\maketitle

\begin{abstract}

This work concentrates on Extremal Regions of Extremum Level (EREL) selection. EREL is a recently proposed feature detector aiming at detecting regions from a set of extremal regions. This is a branching problem derived from segmentation of arterial wall boundaries from Intravascular Ultrasound (IVUS) images. For each IVUS frame, a set of EREL regions is generated to describe the luminal area of human coronary. Each EREL is then fitted by an ellipse to represent the luminal border. The goal is to assign the most appropriate EREL as the lumen. In this work, EREL selection carries out in two rounds. In the first round, the pattern in a set of EREL regions is analyzed and used to generate an approximate luminal region. Then, the two-dimensional (2D) correlation coefficients are computed between this approximate region and each EREL to keep the ones with tightest relevance. In the second round, a compactness measure is calculated for each EREL and its fitted ellipse to guarantee that the resulting EREL has not affected by the common artifacts such as bifurcations, shadows, and side branches. We evaluated the selected ERELs in terms of Hausdorff Distance (HD) and Jaccard Measure (JM) on the train and test set of a publicly available dataset. The results show that our selection strategy outperforms the current state-of-the-art.

\end{abstract}


\section{Introduction}
\label{sec:intro}

This work focuses on segmentation of arterial wall boundaries from Intravascular Ultrasound (IVUS) images where the goal is to develop an automatic way to extract the luminal border of an IVUS image. Various approaches have been studied on this topic such as artificial neural network \cite{su2017artificial}, computational methods using minimization of probabilistic cost function \cite{head1985broyden}, and fast-marching as a representative for most region growing methods \cite{destrempes2014segmentation}\cite{hamdi2013real}\cite{chen2018lumen}. Another state-of-the-art method is known as Extremal Regions of Extremum Level (EREL) \cite{faraji2015extremal,faraji2015erel}, a region detector that is capable of extracting featured pixels from an image. This work is a derivation of \cite{faraji2018segmentation} in the context of Intravascular Ultrasound (IVUS) image segmentation. The region detector EREL is applied on each IVUS frame and several possible EREL regions are produced. Each EREL is fitted by a closest ellipsoid due to the intrinsic shape of coronary artery. Among these fitted ellipsoids, we expect one of them to have the smallest Hausdorff Distance (HD) \cite{balocco2014standardized} to the ground truth luminal border labeled by clinical experts and the goal is to find and assign this ellipse to represent lumen. Based on the pattern observed from the datasets, a possible lumen region is extracted from the last EREL in each IVUS sample and the two-dimensional (2D) correlation coefficients between this region and each EREL is computed. The morphological relationship between each EREL region and its fitted ellipse is studied and used as a compactness measure. These two metrics are combined as a selection strategy to filtrate the EREL with the best performance.

\begin{figure} 
\begin{minipage}[b]{0.5\linewidth}
\centering
\includegraphics[width=1\linewidth]{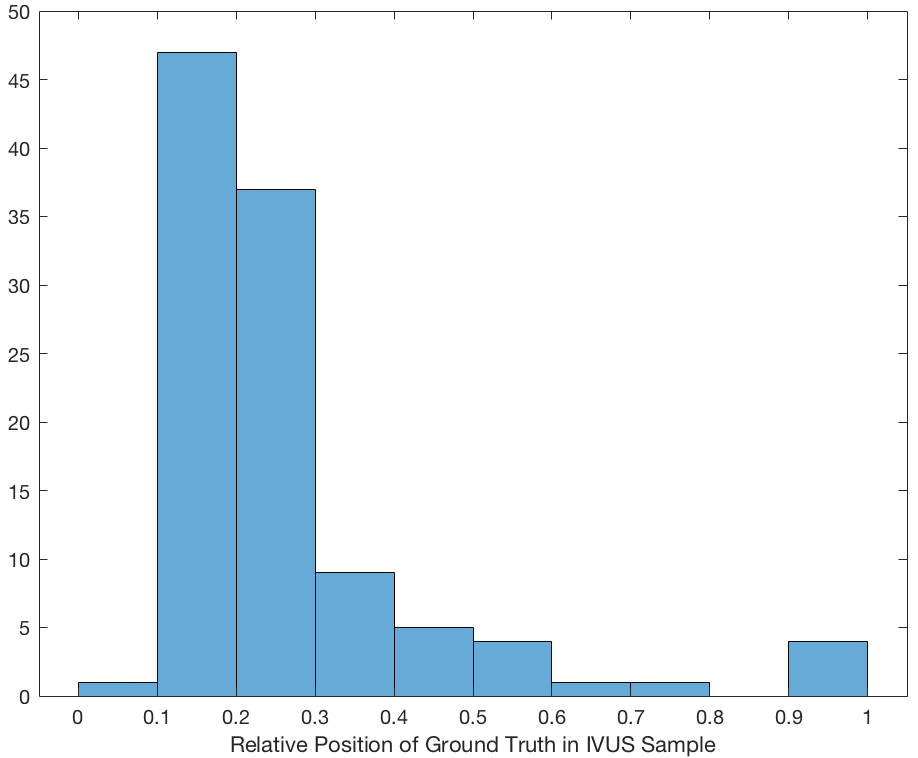} 
\end{minipage}
\begin{minipage}[b]{0.5\linewidth}
\centering
\includegraphics[width=1\linewidth]{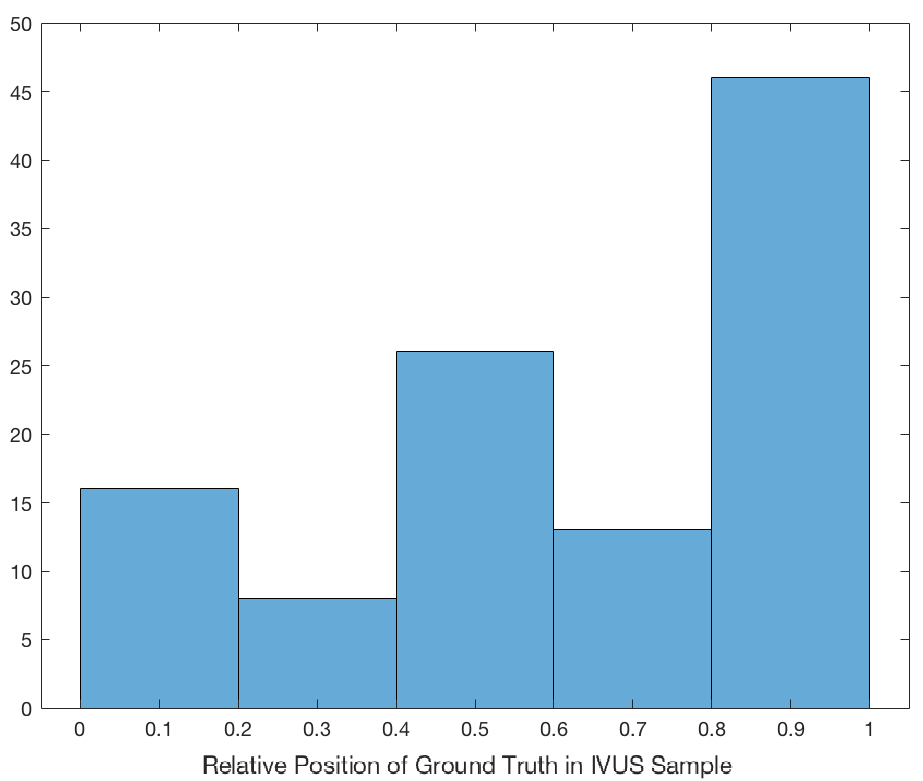}  
\end{minipage}
\label{distribution_ground_truth}
\caption{Distribution of the relative position of ground truth in training and testing set calculated by the index of ground truth divide by the total number of ERELs.} 
\end{figure}

Different from traditional data analysis tasks; in this work, the total number of IVUS frames is 435 and the training set takes only 25.65\% of the total data. Also, the training and testing dataset have a large variation in terms of the number of ERELs and the distribution of ground truth in each sample. In the training set, an IVUS frame has an average of 30 ERELs; whereas in the testing set, an IVUS frame has an average of 7 ERELs. Table~\ref{Comparison} has shown the comparison between both datasets, the distributions of ground truth in both datasets are presented in Fig.~\ref{distribution_ground_truth}. Ground truth ERELs tend to distribute around 10\% to 30\% of the entire data in the training set, whereas there is no evident pattern in the testing set. As a consequence, most feature learning methods such as convolutional neural networks \cite{krizhevsky2012imagenet}, with a high demand on quantity and quality of datasets \cite{flexer1996statistical}, may fail in extracting the relationship between ERELs and will eventually lead to overfit on the training data \cite{reunanen2003overfitting}.

\begin{table}
\caption{Comparison between training dataset and testing dataset \cite{balocco2014standardized}.}
\label{Comparison}
\begin{center}
\begin{tabular}{l l l c}
\textbf{} & \textbf{Training Set}  & \textbf{Testing Set} & \textbf{Overall} \\
\hline
Number of IVUS Frames & 109 & 326 & 435 \\
Percentage in Total Dataset & 25.65\% & 76.71\% & 100\% \\
Best Matching HD (lumen) & 0.19701 & 0.22872 & 0.2260 \\
Total Number of ERELS & 3207 & 2206 & 5413 \\
Average Number of ERELs & 29.61 & 6.77 & 12.7871 \\
Maximum Number of ERELs & 43 & 7 & 43 \\
Minimum Number of ERELs & 6 & 3 & 3 \\
\end{tabular}
\end{center}
\end{table}

In order to deal with this issue, each EREL is treated as an independent entity in the proposed method. The key relationships between ERELs are extracted and the performance of each EREL is evaluated and compared to the others. Based on the variation on the number of ERELs in each sample, we assume there is one and only one EREL detection that possesses the best matching qualifications regardless of the number of competitors. These qualifications are evaluated by computing two morphological metrics for each EREL: the correlation coefficient and the compactness measure. The details are given in the next section. Several advantages of this proposed method include:

\begin{itemize}
\item This method operates immediately on ERELs by extracting them from the original IVUS frames as binary or grayscale regions such that the result is not influenced by other factors where the characteristics of these ERELs can be closely analyzed.
\item This method treats each EREL as an independent entity. Therefore, this method works regardless the number of ERELs in each sample.
\item This method uses two passes where the first pass takes the ERELs with high correlation coefficients and the second pass takes the ERELs with high compactness measures. As a result, accuracy is guaranteed in two folds.
\end{itemize}

\begin{figure} 
	\begin{center}
		\includegraphics[scale=0.35]{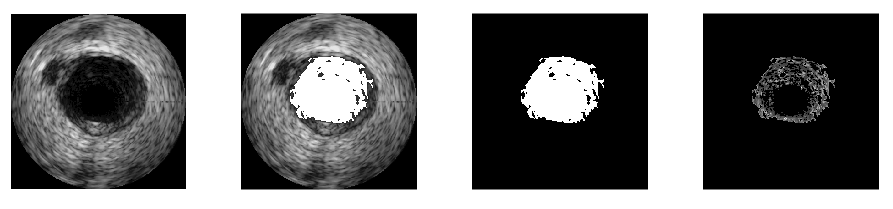}
		\caption{Images from left to right represent: the original IVUS frame, the IVUS frame with EREL region highlighted, the EREL region extracted in binary format, the EREL region extracted in grayscale format, respectively.}
		\label{fig: extractEREL}
	\end{center}
\end{figure}
\section{Method}
\label{sec:method}

The proposed method works based on the pattern observed from EREL regions in each IVUS sample. Some major patterns can be summarized as follows:
\begin{itemize}
\item EREL regions in each sample are ordered in an increasing fashion in terms of the size of detected features. Therefore, features in current EREL are also preserved in the following ones.
\item An EREL feature detector may be influenced by noises such as shadow and bifurcation that lead to irregular shapes of the resulting EREL regions.
\item The ground truth lumen region is relatively stable in terms of the growing trend and also tighter elliptical shapes compared to the others.
\end{itemize}

The procedure in selecting EREL includes several steps. First, EREL regions are preprocessed into binary and grayscale regions respectively. Next, the relationships among ERELs are studied that we process on the last EREL of each sample to extract the possible lumen region. After that, a 2D correlation coefficient is computed between this extracted region and each EREL to only keep the ERELs with high correlations. Then, we further process on these EREL regions with their fitted ellipse to find the EREL with the highest compactness measure as the selected EREL. The details about each step are given as follows.

\begin{figure}
	\begin{center}
		\includegraphics[scale=0.23]{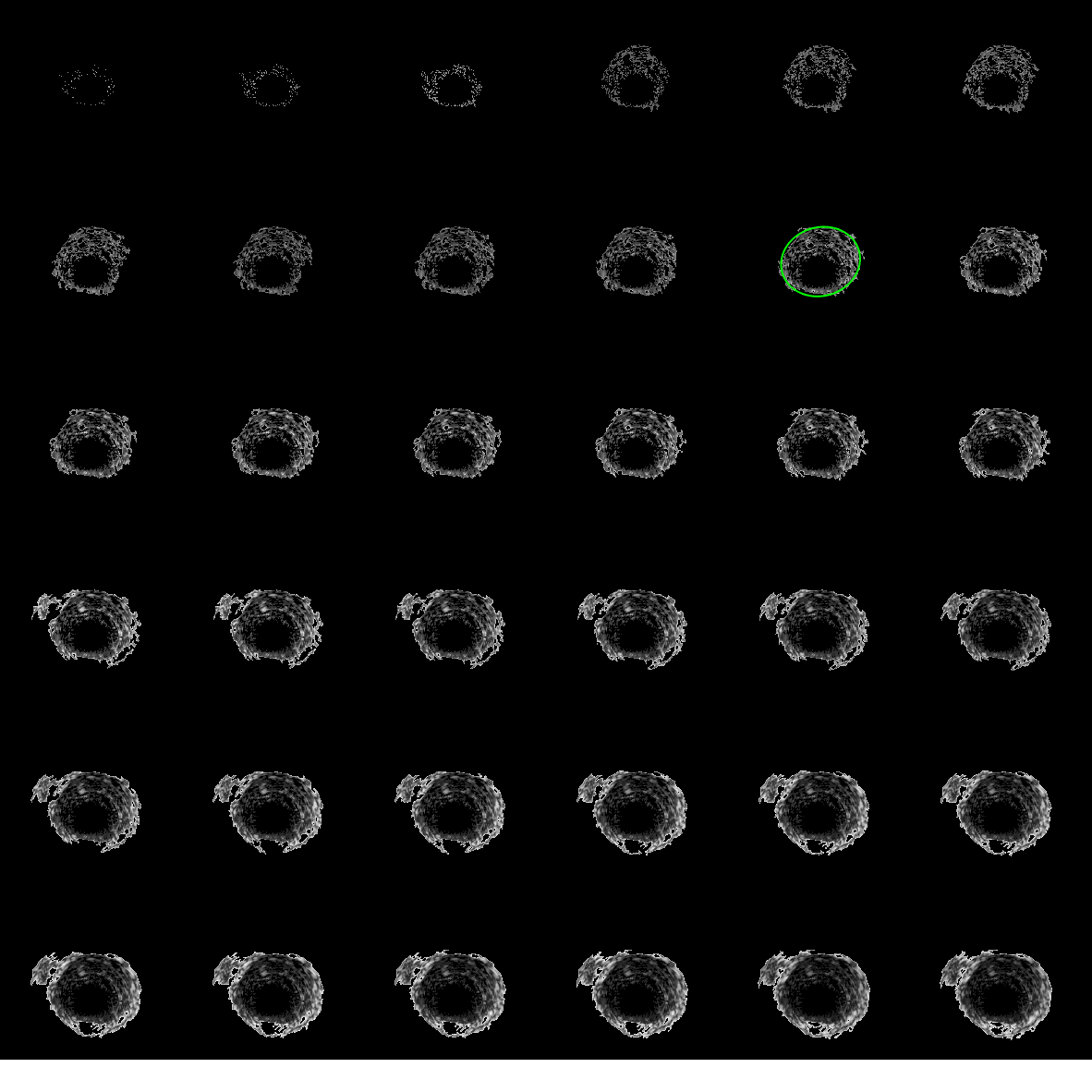}
		\caption{Visualization of EREL regions generated from a sample IVUS frame in dataset. ERELs are visualized in grayscale format, the ground truth EREL is circled in green.}
		\label{fig: intensity36}
	\end{center}
\end{figure}
\subsection{Preprocessing}
In the dataset, each EREL is represented by a set of pixel coordinates. In order to analyze their characteristics and relationships, these ERELs are extracted in binary form and grayscale form respectively, shown in Fig.~\ref{fig: extractEREL}.
\begin{figure} 
	\begin{center}
		\begin{tabular}{m{0.5cm} m{6cm}}
			(a)&\includegraphics[scale=0.15]{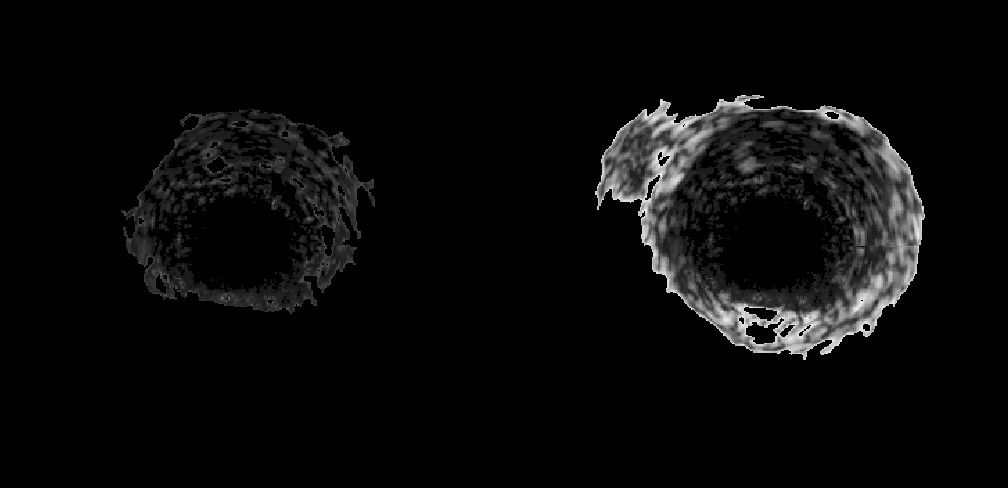} \\
			(b)&\includegraphics[scale=0.2]{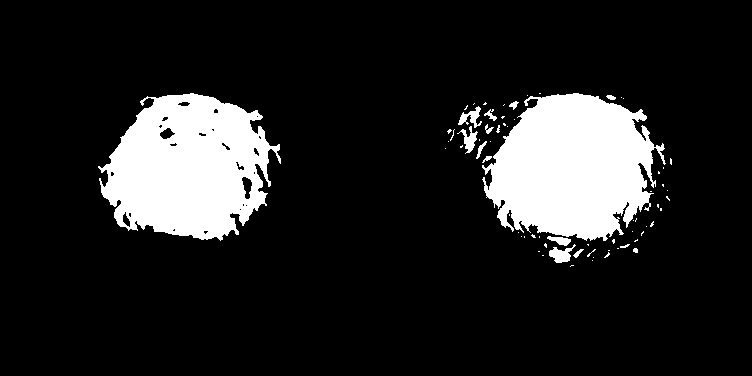} \\
		\end{tabular}
		
		\caption{(a) Comparison between the ground truth (left) and the last EREL (right). (b) Comparison between the ground truth (left) and the approximate lumen region (right) in binary format.}
		\label{fig: comparison1}
	\end{center}
\end{figure}

\subsection{Evaluation on Correlation}
Fig.~\ref{fig: intensity36} visualizes EREL regions for an IVUS sample in grayscale form, where the ground truth EREL is circled. As we observe into these ERELs, we can see that the general shape of the ground truth EREL along with its intensity are preserved in all the subsequent ERELs. A direct comparison between the ground truth EREL and the last EREL in this sample is shown in Fig.~\ref{fig: comparison1}(a). In this last EREL, the low intensity pixels in the middle preserve the lumen region and the idea is to process on this frame to extract the approximate lumen region. A 2D correlation coefficient $r$ is computed for each EREL $A$ and this approximate lumen region $B$:
\begin{figure}  
	\begin{minipage}[b]{0.5\linewidth}
		\centering
		\includegraphics[width=1\linewidth]{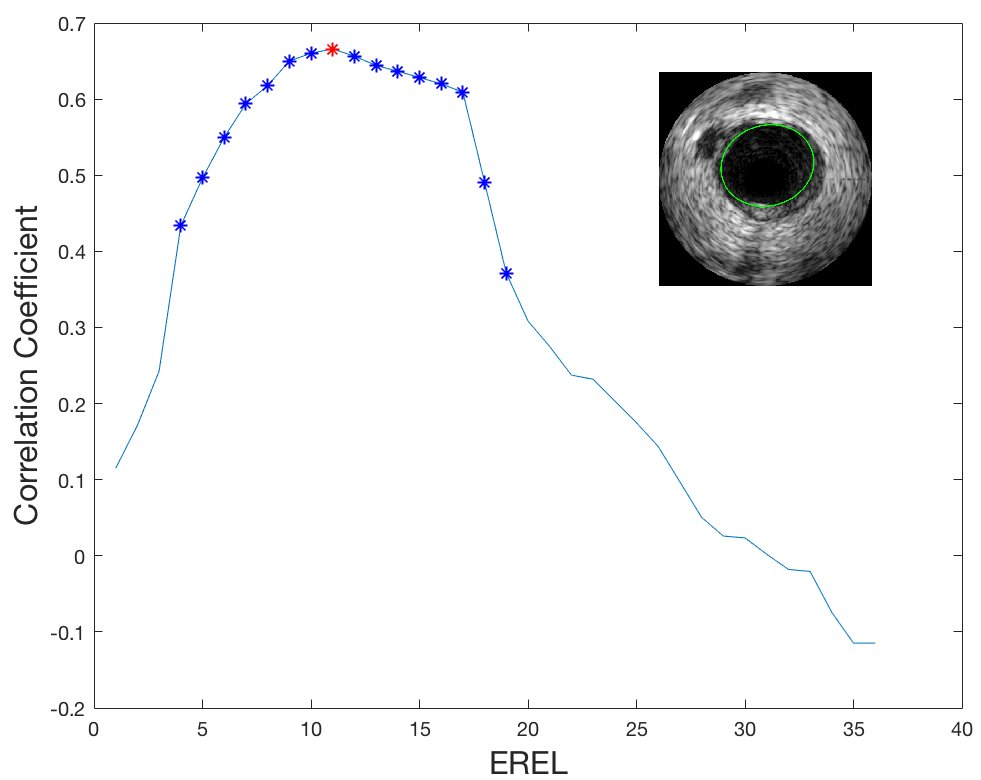} 
	\end{minipage}
	\begin{minipage}[b]{0.5\linewidth}
		\centering
		\includegraphics[width=1\linewidth]{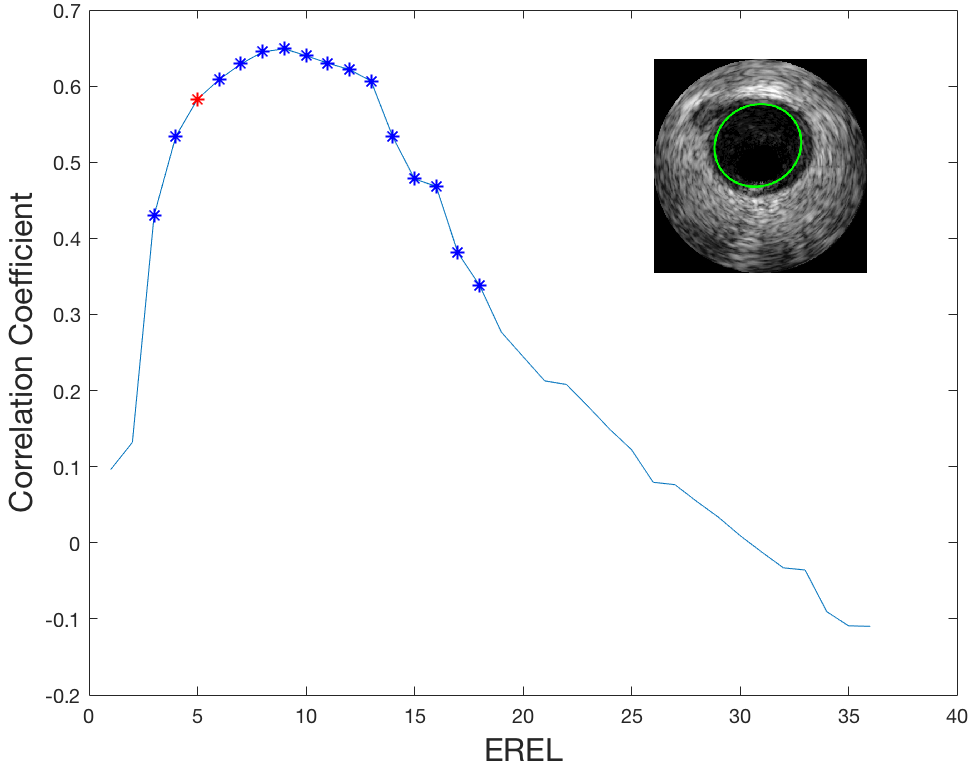}  
	\end{minipage} 
	\begin{minipage}[b]{0.5\linewidth}
		\centering
		\includegraphics[width=1\linewidth]{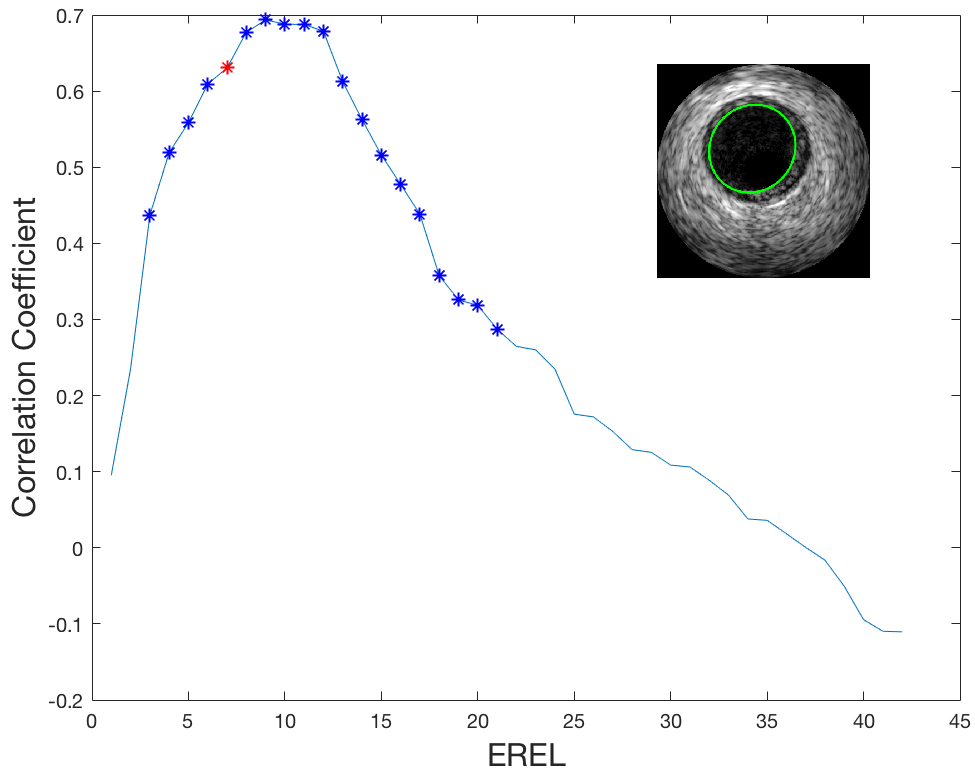}
	\end{minipage} 
	\begin{minipage}[b]{0.5\linewidth}
		\centering
		\includegraphics[width=1\linewidth]{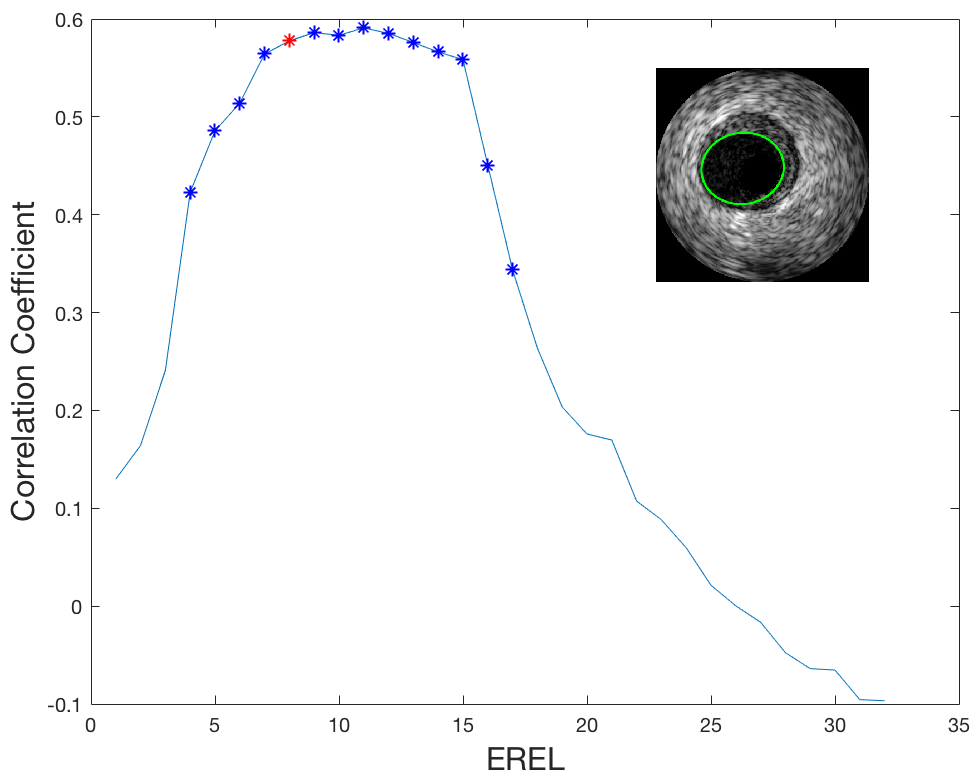} 
	\end{minipage} 
	
	\caption{Correlation plots of sample IVUS frames where the correlation coefficients are computed using Equation~\ref{equation1}. IVUS frames are presented in top-right corner of each plot with ground truth contour highlight in green. Blue markers on each plot represent those selected ERELs with correlation coefficients higher than the threshold. The red marker on each plot represents the ground truth EREL.}
	\label{correlation_plot}
\end{figure}

\begin{equation} \label{equation1}
r = \frac{\sum\limits_{m}\sum\limits_{n}(A_{mn} - \bar{A})(B_{mn} - \bar{B})}{\sqrt[]{(\sum\limits_{m}\sum\limits_{n}(A_{mn} - \bar{A})^2)(\sum\limits_{m}\sum\limits_{n}(B_{mn} - \bar{B})^2)}} 
\end{equation}

where $\bar{A}$ is the average intensity of $A$, $\bar{B}$ is the average intensity of $B$ and $m, n$ represent columns and rows of $A, B$ respectively. Presumably, the ground truth should have a high correlation with this region.

Therefore, the average intensity of the last EREL is calculated as a threshold. we process on the last EREL to keep only the low intensity pixels and the result is shown in Fig.~\ref{fig: comparison1}(b). The region on the left is the ground truth region in binary form and the region on the right is the extracted region. As we can see these two regions are close and they should have a high correlation.
\begin{figure}
	\begin{center}
		\includegraphics[scale=0.27]{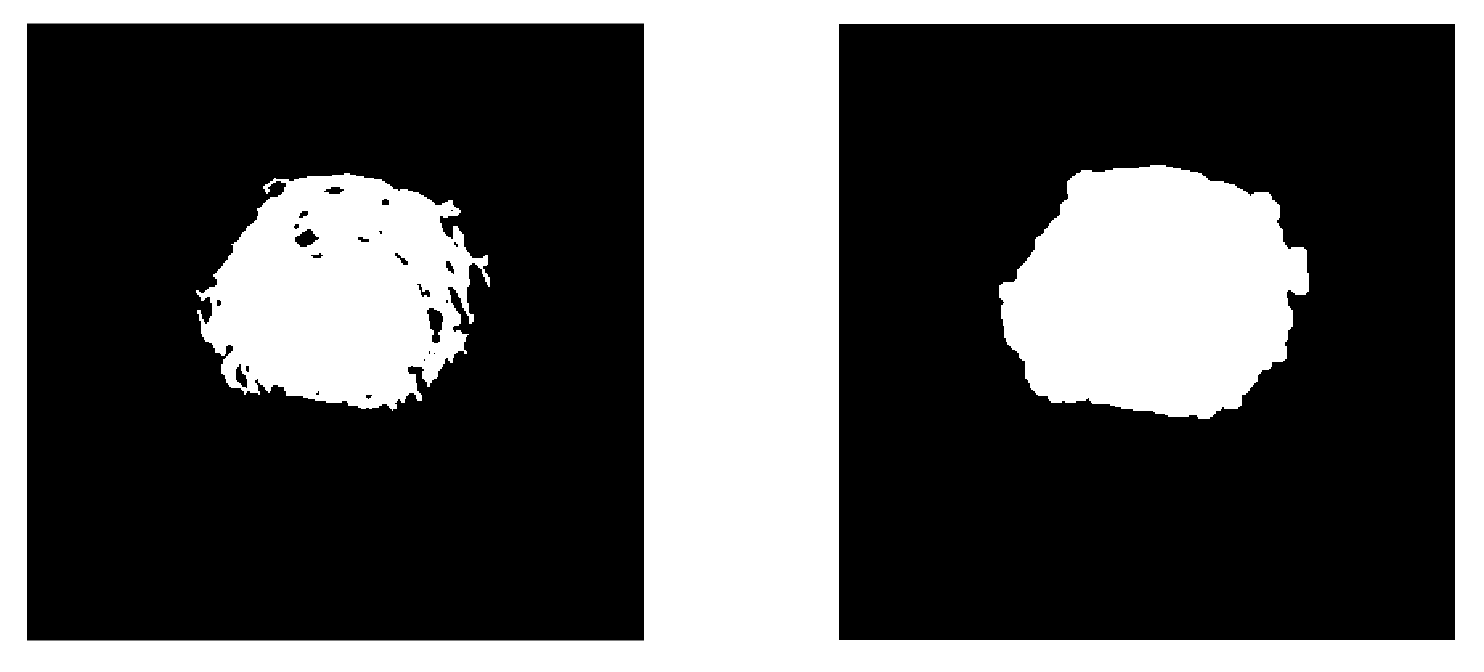}
		\caption{EREL region before (left) and after (right) dilation using disk-shaped structuring element of radius = 6.}
		\label{fig: dilation}
	\end{center}
\end{figure}
\begin{figure}
	\begin{center}
		\includegraphics[scale=0.35]{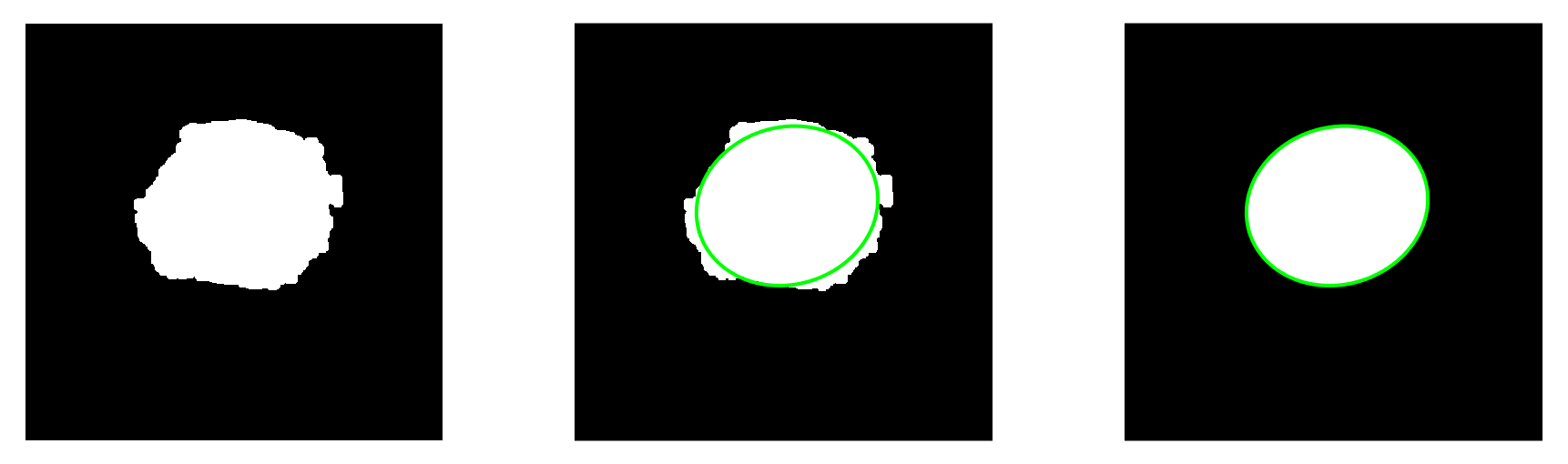}
		\caption{Illustration of masking an EREL region. Images from left to right represent: the EREL region after dilation, the EREL region and its fitted ellipse in green, intersected area of EREL and its fitted ellipse, respectively.}
		\label{fig: mask}
	\end{center}
\end{figure}
Some sample plots for correlation are shown in Fig.~\ref{correlation_plot}. It can be observed that these plots tend to be right-skewed \cite{simon1960some} and the ground truth ERELs (marked in red) are located around the peak of each plot. Without missing any ERELs, we are setting the average of these correlations as a threshold, and output all the ERELs with correlations higher than this threshold. Hence, the resulting list of ERELs is guaranteed to contain the ground truth EREL.

\subsection{Evaluation on Compactness}
An EREL region is eventually defined by its contour; however, they may subject to internal cracks since ERELs in their nature are regions selected from a set of extremal points \cite{faraji2015extremal,faraji2015erel}. Therefore, a morphological dilate operation is performed on each EREL. This dilation ensures that some insignificant cracks are filled to some extent, and gaps between ERELs caused by bifurcations can be narrowed where the general shape of an EREL is still preserved (shown in Fig.~\ref{fig: dilation}).

Due to the intrinsic shape of a luminal border, each EREL region is described by an elliptical contour \cite{faraji2018segmentation}. The intuitive idea behind this step is that among all the EREL regions, the best matching EREL should have the tightest relationship with the fitted ellipse. Therefore, the ellipse is used as a boolean mask and the intersection between the EREL region and the ellipse is extracted (shown in Fig.~\ref{fig: mask}).

\begin{figure}  
	\begin{minipage}[b]{0.5\linewidth}
		\centering
		\includegraphics[width=1\linewidth]{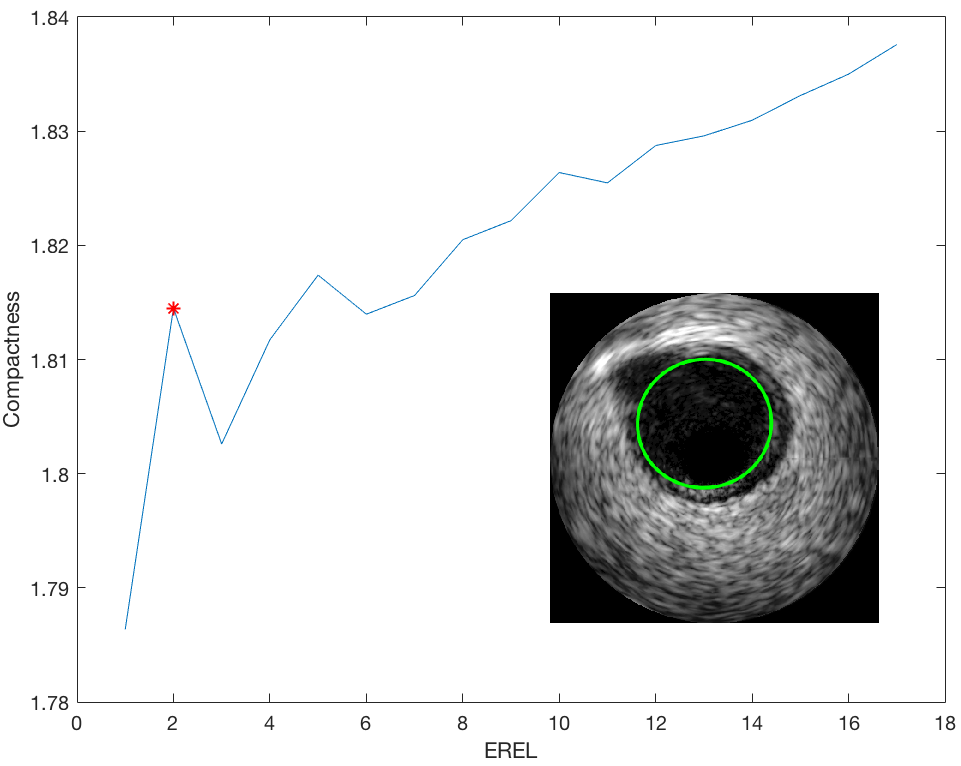} 
	\end{minipage}
	\begin{minipage}[b]{0.5\linewidth}
		\centering
		\includegraphics[width=1\linewidth]{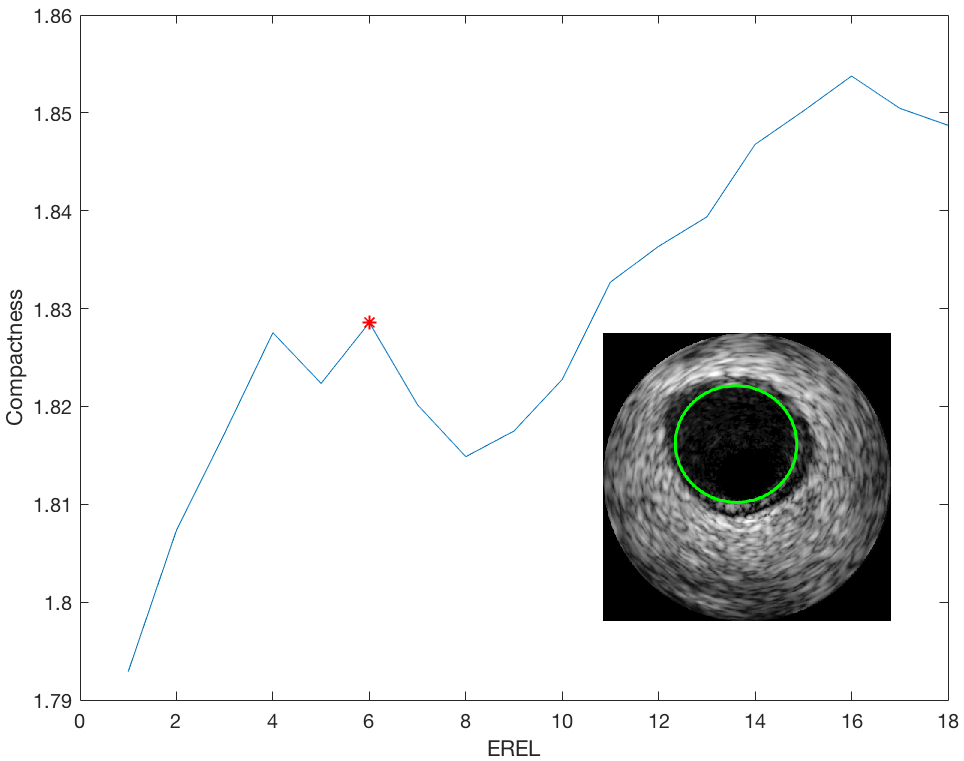}  
	\end{minipage} 
	\begin{minipage}[b]{0.5\linewidth}
		\centering
		\includegraphics[width=1\linewidth]{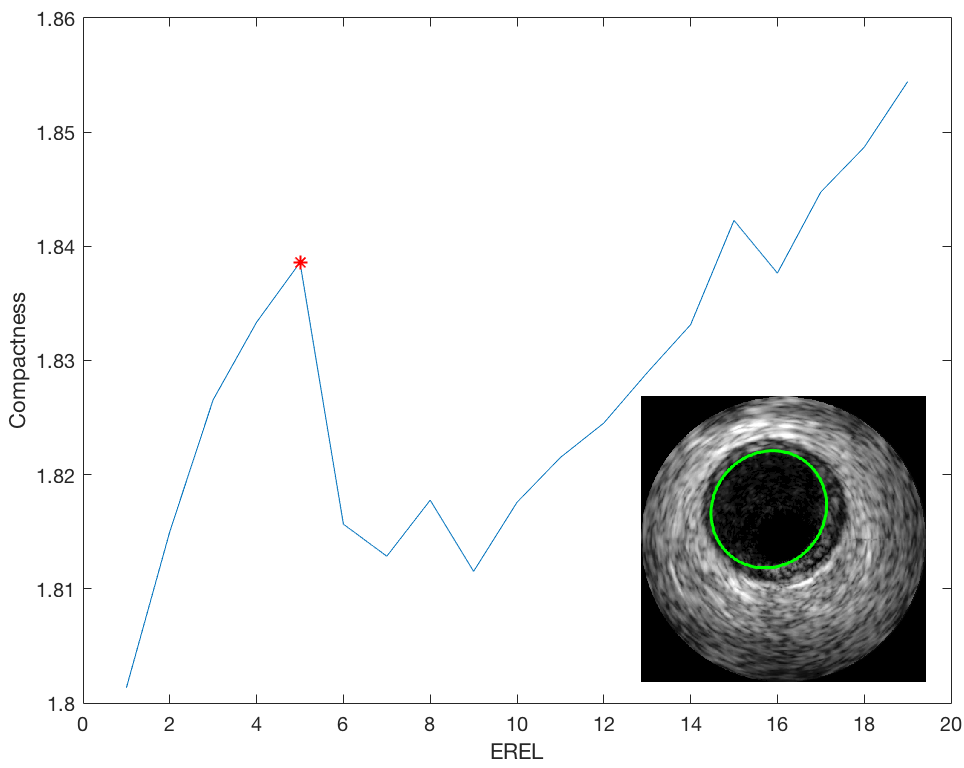}
	\end{minipage} 
	\begin{minipage}[b]{0.5\linewidth}
		\centering
		\includegraphics[width=1\linewidth]{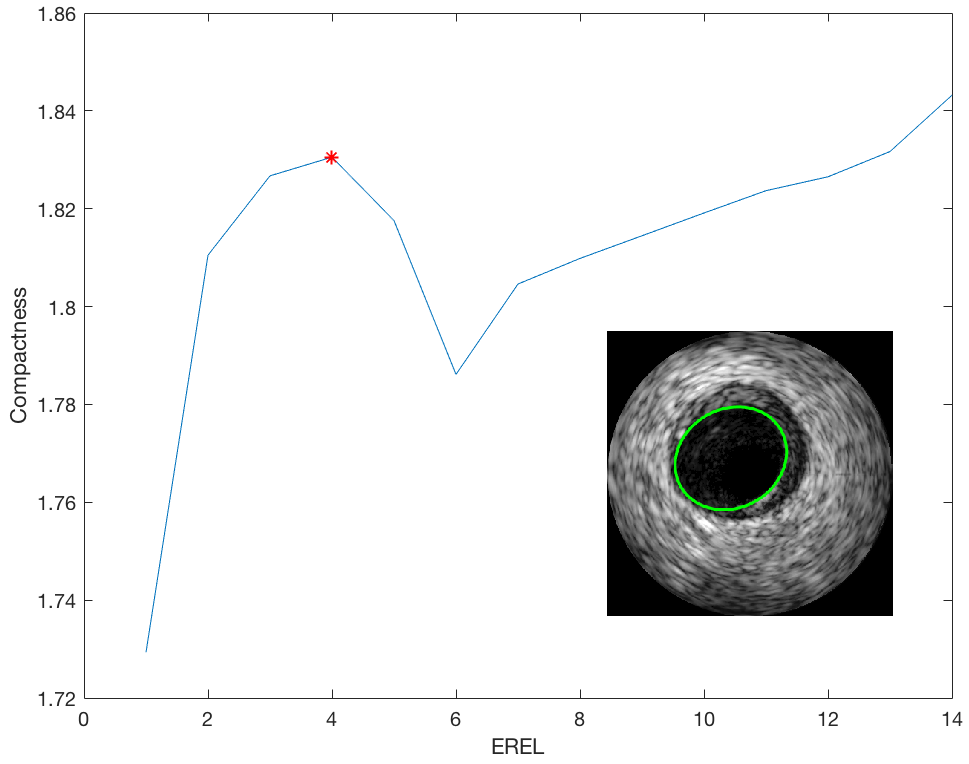} 
	\end{minipage} 
	\caption{Compactness plots of sample IVUS frames where the compactness values are computed as $M1 + M2$ where $M1, M2$ are calculated from Equation~\ref{equationM1} and Equation~\ref{equationM2}. IVUS frames are presented in bottom-right corner of each plot with ground truth contour highlight in green. The red marker on each plot represents the ground truth EREL in the corresponding IVUS sample.}
	\label{compactness_plot}
\end{figure}

For each EREL region, two measurements will be calculated:
\begin{enumerate}
\item The intersection between EREL and its fitted ellipse over the fitted ellipse:

\begin{equation} \label{equationM1}
M_1 = \frac{Area_{ellipse} \cap Area_{EREL}}{Area_{ellipse}}
\end{equation}

This measurement guarantees a minimal amount of missing pixels within the fitted elliptical region after the dilation operation. Especially, it can effectively screen out those ERELs with shadow and bifurcation inaccurately included within the detected area.

\item And the intersection between EREL and its fitted ellipse over the whole EREL region:
\begin{equation} \label{equationM2}
M_2 = \frac{Area_{ellipse} \cap Area_{EREL}}{Area_{EREL}}
\end{equation}

This measurement guarantees that most of the detected pixels are fitted into the ellipse. It can also effectively exclude those EREL regions with bifurcation and shadow outside of the fitted ellipse that cause irregular boundary of the EREL.

\end{enumerate}

Then these two measurements, $M_1$ and $M_2$, are summed and used together as a compactness standard to define the suitability of an EREL region. A qualified EREL region is expected to have a higher value in this metric than the others in the same IVUS frame sample. Later, these measurements are arranged with the same ordering as the ERELs and the next subsection will describe how the best matching EREL is extracted.

\subsection{EREL Selection}

Fig.~\ref{compactness_plot} visualizes the compactness plots for some sample frames, where the ground truth ERELs are marked in red. We can conclude that these plots obey a similar growing trend and the best matching EREL locates at one of the local maxima of each plot. It is also worth noting that other local maxima may correspond to the shadowed area within the lumen, the media border, and also some interferential regions with conformance to elliptical shape.

Without loss of generality, for IVUS frames where several local maxima exist, the best matching EREL is selected from the first few local maxima where they have a high possibility to represent lumen. For samples with small number of ERELs where the compactness measure continuously grow and local maximum does not exist, the global maximal EREL is chosen. The result of this method is presented and analyzed in the next section.

\section{Experimental Results}
\label{sec:results}

Among all the EREL regions, the one with the shortest Hausdorff Distance (HD) to the hand-annotated ground truth is taken as the golden standard. These numbers represent the best result that an algorithm can achieve in this EREL selection task. The performance of the proposed method is summarized in Table~\ref{table_performance_HD} evaluated under five categories: the general performance and the performances on frames with no artifacts, with bifurcation, with side vessels and with shadow. A comparison between the proposed method and the result in original work \cite{faraji2018segmentation} is shown in Table~\ref{table_comparison_HD} where the proposed method outperforms the original method especially when no artifacts present.   

\begin{table}
\begin{center}\scriptsize
\caption{Performance of the proposed method in lumen detection. The metrics being used are Hausdorff Distance (HD) and Jaccard Measure (JM) evaluated in average and standard deviation. Performance is evaluated under five dataset categories: the general performance, frames with no artifacts, frames with bifurcation, frames with side vessels and frame with shadow. GT denotes the reported Ground Truth.}
\label{table_performance_HD}	
\begin{tabular}{p{0.15\textwidth} p{0.1\textwidth} p{0.15\textwidth} p{0.2\textwidth} p{0.15\textwidth} p{0.2\textwidth}}
\textbf{} & \textbf{Dataset} & \textbf{HD} (GT)  & \textbf{HD} (Proposed)&\textbf{JM} (GT)  & \textbf{JM}(Proposed)\\
\hline
General & Training  &  0.1970 (0.09) & {0.3159 (0.17)} &  0.9123 (0.03) & {0.8761 (0.07)} \\
        &  Testing &  0.2287 (0.14) & {0.2952 (0.24)} &  0.8906 (0.06) & {0.8747 (0.07)} \\
\\
No Artifact &Training &  0.1861 (0.06) & {0.3080 (0.16)}&  0.9138 (0.03) & {0.8755 (0.07)} \\
            &Testing &  0.2076 (0.16) & {0.2771 (0.25)}&  0.8978 (0.06) & {0.8864 (0.06)} \\
\\
Bifurcation &  Training & 0.2490 (0.19) & {0.3805 (0.22)}&  0.9021 (0.05) & {0.8706 (0.06)} \\
            &Testing &  0.4230 (0.10) & {0.5544 (0.19)}&  0.8854 (0.04) & {0.7791 (0.11)} \\
\\
Side Vessels &  Training &  0.2426 (0.00) & {0.2426 (0.00)}&  0.9269 (0.00) & {0.9269 (0.00)} \\
             &  Testing &  0.1914 (0.15) & {0.2406 (0.24)}&  0.8809 (0.06) & {0.8872 (0.06)} \\
\\
Shadow &  Training &  0.1987 (0.07) & {0.2848 (0.13)} &  0.9153 (0.02) & {0.8851 (0.05)} \\
       &  Testing &  0.2333 (0.12) & {0.2906 (0.22)} &  0.8835 (0.04) & {0.8591 (0.08)}  \\
\end{tabular}
\end{center}
\end{table}

\begin{table}[!ht]
\begin{center}
\caption{Comparison between the state-of-the-art \cite{faraji2018segmentation} and the proposed method in lumen detection. The metrics being used are Hausdorff Distance (HD) and Jaccard Measure (JM), performance is evaluated in average and standard derivation.}
\label{table_comparison_HD}
\begin{tabular}{p{0.15\textwidth} m{0.15\textwidth} m{0.2\textwidth} m{0.15\textwidth} m{0.2\textwidth} }
\textbf{} & \textbf{HD} \cite{faraji2018segmentation} & \textbf{HD} (Proposed) & \textbf{JM} \cite{faraji2018segmentation} & \textbf{JM} (Proposed) \\
\hline
General &  0.30 (0.20) & {0.2952 (0.24)} & 0.87 (0.06) & {0.8747 (0.07)} \\
No Artifact &  0.29 (0.17) & {0.2771 (0.25)}& 0.88 (0.05) & {0.8864 (0.06)}\\
Bifurcation &  0.53 (0.34) & {0.5544 (0.19)}&0.79 (0.12) & {0.7791 (0.11)} \\
Side Vessels &  0.24 (0.11) & {0.2406 (0.24)}&0.87 (0.05) & {0.8872 (0.06)} \\
Shadow &  0.29 (0.20) & {0.2906 (0.22)}&0.86 (0.07) & {0.8591 (0.08)} \\
\end{tabular}
\end{center}
\end{table}

\section{Conclusion}\label{sec::conclusion}

This method uses the morphological characteristics of EREL regions as metrics to evaluate these ERELs. Other feature learning methods may require pixel-to-pixel comparison between every two EREL regions and would result in high computational consumption; where this proposed method operates on independent ERELs which makes it fast and straightforward. 

Nevertheless, there are still some improvements needed for this method. In the first pass, this method extracts a possible lumen region from the last EREL in each series and computes the correlation coefficient between each EREL and this extracted region. For a small IVUS sample where the last EREL is similar to all the other ones, this method fails to extract a possible lumen region which may lead to negative correlation in all comparisons. The second pass works on the basis of the compactness of an EREL region and its fitted ellipse. A major issue comes from this method is that any EREL conformed to elliptical shape may be wrongly chosen as a suitable representation. Therefore, incorporating the characteristics of lumen into this method is necessary and beneficial to eliminate these interferences.

Furthermore, considering the simplicity and power of this combined correlation and compactness evaluations, it is also likely to produce a better result if the these metrics can be used as supplementary tools in other EREL selection algorithms to screen out the inappropriate ERELs.

The original study also involves assigning an EREL as media region of human coronary where in this work we focus on lumen only. EREL selection as media adopts a similar idea with some variations due to the particular characteristics of media, this task will be assigned as the future work of this study.

\bibliographystyle{splncs}
\bibliography{refs}
\end{document}